\pdfoutput=1

\documentclass[10pt,twocolumn,letterpaper]{article}

\usepackage[pagenumbers]{cvpr} % To force page numbers, e.g. for an arXiv version

\usepackage[dvipsnames]{xcolor}

\usepackage{multirow}
\usepackage{caption}
\usepackage{subcaption}
\usepackage{tikzscale}
\usepackage{pgfplots,pgfplotstable}

\definecolor{cvprblue}{rgb}{0.21,0.49,0.74}
\usepackage[pagebackref,breaklinks,colorlinks,citecolor=cvprblue]{hyperref}

\def\paperID{*****} % *** Enter the Paper ID here
\def\confName{CVPR}
\def\confYear{2024}

\title{CoBEVFusion: Cooperative Perception with LiDAR-Camera Bird's-Eye View Fusion}

\author{Donghao Qiao, Farhana Zulkernine\\
Queen’s University, Canada\\
{\tt\small \{d.qiao, farhana.zulkernine\}@queensu.ca}
}

\begin{document}
\maketitle
\begin{abstract}
Autonomous Vehicles (AVs) use multiple sensors to gather information about their surroundings. By sharing sensor data between Connected Autonomous Vehicles (CAVs), the safety and reliability of these vehicles can be improved through a concept known as cooperative perception. However, recent approaches in cooperative perception only share single sensor information such as cameras or LiDAR. In this research, we explore the fusion of multiple sensor data sources and present a framework, called CoBEVFusion, that fuses LiDAR and camera data to create a Bird's-Eye View (BEV) representation. The CAVs process the multi-modal data locally and utilize a Dual Window-based Cross-Attention (DWCA) module to fuse the LiDAR and camera features into a unified BEV representation. The fused BEV feature maps are shared among the CAVs, and a 3D Convolutional Neural Network is applied to aggregate the features from the CAVs. Our CoBEVFusion framework was evaluated on the cooperative perception dataset OPV2V for two perception tasks: BEV semantic segmentation and 3D object detection. The results show that our DWCA LiDAR-camera fusion model outperforms perception models with single-modal data and state-of-the-art BEV fusion models. Our overall cooperative perception architecture, CoBEVFusion, also achieves comparable performance with other cooperative perception models.

\end{abstract}    
\section{Introduction}
\label{sec:intro}

Light Detection and Ranging (LiDAR) and camera are crucial sensors in Autonomous Vehicles (AVs) for perceiving surrounding traffic information. Cameras provide rich color and texture information, making them ideal for detecting small objects, identifying traffic signs, and recognizing lanes. LiDAR, on the other hand, scans the surroundings and provides a 3D view with precise distance measurements of objects. Despite their strengths, both sensors have limitations due to their inherent characteristics. For example, cameras are sensitive to light and lack distance information, while LiDARs lack color and texture information and can be affected by severe weather conditions. Combination of two sensors can overcome the limitations of a single sensor and provide a more comprehensive understanding of the traffic environment for perception.

\begin{figure}[t]
\centering
\includegraphics[width=0.95\columnwidth]{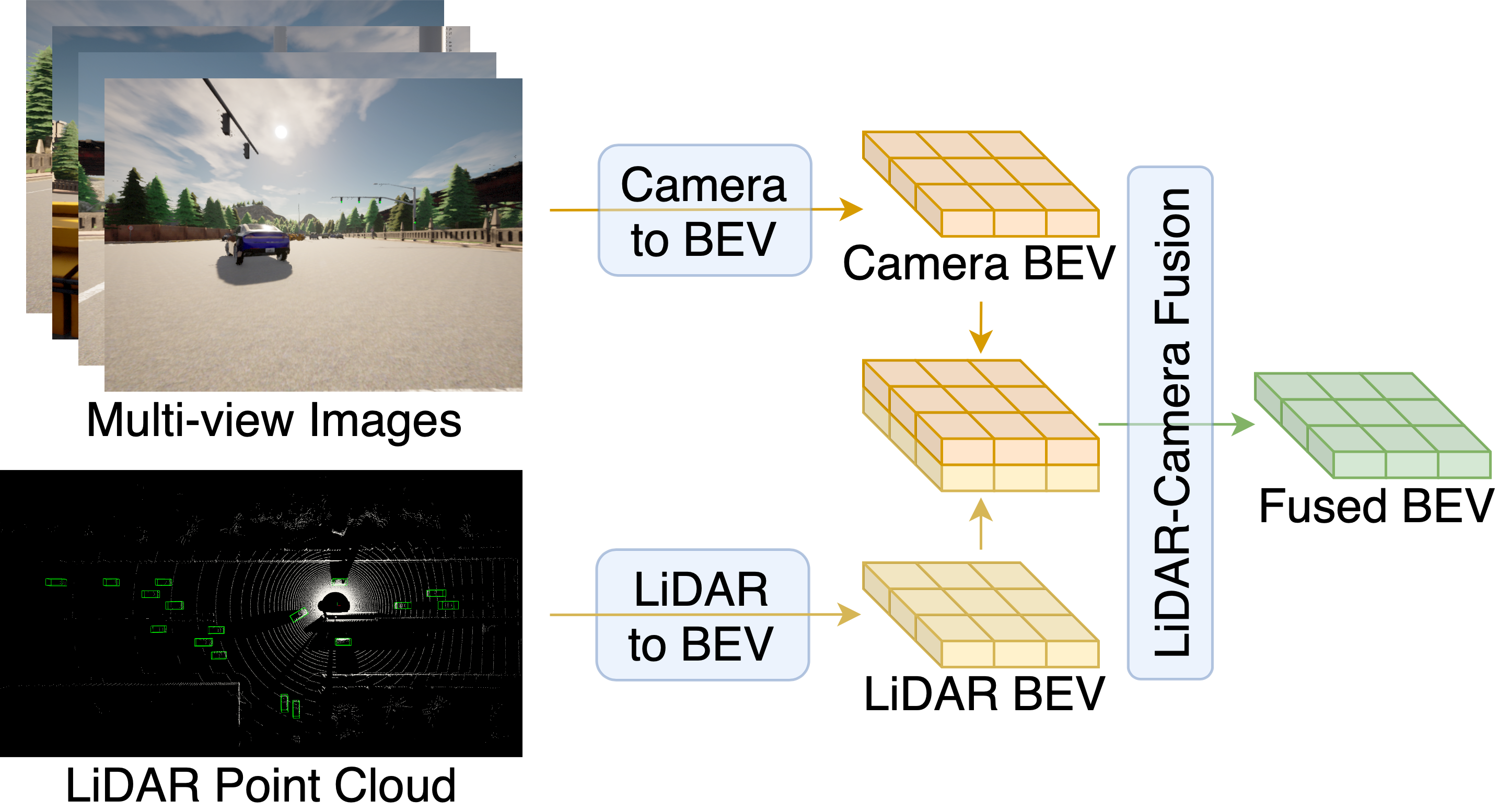}
\caption{Architecture of the LiDAR-camera Bird's-Eye View (BEV) fusion.}
\label{fig:single}
\end{figure}

The pioneering works by~\cite{mv3d,avod} present proposal-level multi-modal sensor fusion, which utilizes the image features to help generate 3D object detection proposals. These models are limited in their ability to utilize contextual information. Other works~\cite{pointpainting,pointaugmenting,fusionpainting} augment point cloud with image features generated by semantic segmentation network or 2D object detection network for 3D object detection. 
%The point cloud augmentation models %\cite{pointpainting,fusionpainting,pointaugmenting} require pretrained CNN models to %augment LiDAR point cloud and then process the augmented point cloud for further prediction. 
However, these point cloud augmentation models have low parallelism, leading to longer prediction times and inability to fully utilize the information from images due to the sparsity of the point cloud. Some works~\cite{mv3d,avod,deepfusion,pointpainting,fusionpainting} only consider the single front-view perception, whereas current AVs are equipped with multiple cameras to capture the surrounding traffic. Some researchers propose LiDAR-camera fusion with multi-view camera images~\cite{futr3d,bevfusion2,bevfusion1}. When fusing multi-view images and LiDAR point clouds, it is critical to unify the feature representation to diminish the loss of geological and characteristic information during the feature fusion. 

The BEV representation is flexible and feasible in tackling multiple perception tasks including semantic segmentation~\cite{cobevt,cvt} and 3D object detection~\cite{pointpillars,voxelnet}. As shown in Fig.~\ref{fig:single}, we propose an end-to-end LiDAR point cloud and camera images fusion framework to utilize the intermediate BEV representations. The multi-view camera images and the LiDAR point cloud are encoded and projected into BEV. The biggest challenge of feature fusion is how to effectively and fully utilize the features from the two data sources. During data processing, due to the effects of resolution and topology, the BEV representations of LiDAR and camera need further alignment. Channel selection network~\cite{bevfusion2} only selects the information from the channel, but cannot align the two feature maps. The kernel size of CNN limits the receptive field in feature extraction and alignment~\cite{bevfusion1}. The attention mechanism in transformer is able to have a global receptive field which can be utilized for multi-modal feature fusion. Cross-attention can be used to augment one kind of data feature~\cite{deepfusion}, however, both LiDAR and camera have their own advantages which are suitable for different perception tasks. LiDAR provides precise distance information for the objects with longer detection distance than camera. On the other hand, camera captures RGB images containing texture and semantic information, which is more suitable for traffic scene segmentation.

In this research, we propose a novel cross-attention module called Dual Window-based Cross-Attention (DWCA) that consists of two cross-attention blocks with reversed inputs of both LiDAR and camera features. This LiDAR-camera fusion network aligns and fuses the LiDAR and camera BEV representations, and generates a fused BEV representation for perception. Two cross-attention blocks make the LiDAR and camera reinforce each other, thus the overall module can adapt to different perception tasks.

Despite using multi-sensor fusion, the single vehicle perception still faces various challenges and limitations caused by occlusion, blind spots and limited resolution. Cooperative perception can compensate for the limitations of the traditional single vehicle perception. Based on the LiDAR-camera BEV fusion framework, we develop a cooperative perception framework to address the above limitations of the single vehicle perception. Connected Autonomous Vehicles (CAVs) are equipped with Vehicular Communication (VC) systems, which allow CAVs to communicate with other CAVs~\cite{qiao2023adaptive,opv2v} or roadside infrastructures~\cite{infracooper,v2xvit,dairv2x}, and share traffic information within a limited communication range. The information can be the sensor data from both camera~\cite{cobevt} and LiDAR~\cite{qiao2023adaptive,v2xvit,opv2v} or the fused multi-modal representations. CAVs can aggregate the information received from the other CAVs with their own data to improve the accuracy of perception, and enhance the robustness of the perception system as well as the safety of AVs.

Data sharing and fusion in cooperative perception can be split into three classes: early fusion, intermediate fusion and late fusion. Early fusion shares large amount of raw sensor data which contains full contextual information, but uses high bandwidth. Late fusion shares predicted outputs which require low bandwidth, but contain no contextual information. In this research, we use intermediate fusion by sharing the fused LiDAR-camera BEV representations that contain contextual information and occupies less bandwidth than the original sensor data. 

\textbf{Contribution.} The contributions of this work can be summarized as follows:
\begin{itemize}[noitemsep]
    \item We propose a novel Dual Window-based Cross-Attention (DWCA) model for LiDAR-camera BEV fusion.
    \item We develop a cooperative perception framework, CoBEVFusion, which enables perception of multi-modal sensor data by CAVs.
    \item The proposed approach is validated on a large-scale cooperative perception benchmark dataset OPV2V~\cite{opv2v} with two perception tasks namely BEV semantic segmentation and 3D object detection.
    \item We compare our model with single vehicle perception models with single and multiple data modalities. We also compare our CoBEVFusion with SOTA cooperative perception models.
\end{itemize}

The experiments demonstrate that an effective LiDAR-camera fusion model can improve the perception accuracy. Our proposed DWCA surpasses the perception models with single-modal data~\cite{pointpillars,cobevt,cvt} and the SOTA BEV fusion models~\cite{bevfusion2,bevfusion1}. Our cooperative perception architecture, CoBEVFusion, utilizes the fused LiDAR-camera representation, and outperforms the single vehicle perception models and most SOTA cooperative perception models~\cite{fcooper,v2vnet,cobevt,opv2v} on the OPV2V BEV semantic segmentation and 3D object detection tasks.

The rest of the paper is organized as follows. Section~\ref{sec:related_work} describes the related work on cooperative perception and feature fusion models. Our proposed cooperative perception framework and feature fusion models are illustrated in Section~\ref{sec:methodology}. The experimental results and our discussion are presented in Section~\ref{sec:experiments}. Section~\ref{sec:conclusion} concludes the paper with a list of future work.
\section{Related Work}
\label{sec:related_work}

In this section, we survey the related work on multi-modal sensor fusion with a focus on LiDAR-camera fusion models, multi-view image processing in camera streams, and cooperative perception models.

\subsection{Multi-Modal Sensor Fusion}
Several approaches have been proposed for 3D object detection, including Multi-View 3D Object Detection (MV3D)~\cite{mv3d} and Aggregate View Object Detection (AVOD)~\cite{avod}. These methods fuse proposals generated from both image and point cloud representations. PointPainting~\cite{pointpainting} and FusionPainting~\cite{fusionpainting} incorporate semantic segmentation information from images to enhance the point clouds. PointAugmenting~\cite{pointaugmenting}, on the other hand, enhances the LiDAR point clouds with features generated by a 2D object detection network.
% Multi-View 3D Object Detection (MV3D)~\cite{mv3d} and Aggregate View Object Detection (AVOD)~\cite{avod} are proposed for 3D object detection and they fuse the proposals generated by using image and point cloud representations. PointPainting~\cite{pointpainting} and FusionPainting~\cite{fusionpainting} utilize the semantic segmentation information from images to enrich the point clouds. PointAugmenting~\cite{pointaugmenting}, on the other hand, augments the LiDAR point clouds with the image features generated by 2D object detection network. 
DeepFusion~\cite{deepfusion} utilizes cross-attention to align the LiDAR and camera feature representations during the fusion process. TransFusion~\cite{transfusion} condenses the image features along the vertical dimension and then projects features onto the BEV plane using cross-attention to fuse with the LiDAR BEV feature. BEVFusion~\cite{bevfusion2,bevfusion1} projects multi-view images into BEV using a modified version of Lift-Splat~\cite{lss}. \cite{bevfusion1} utilizes a simple 2D CNN for feature alignment, and \cite{bevfusion2} proposes dynamic fusion, which is a channel-wise feature selection network.

In order to avoid using the expensive LiDAR, Simple-BEV~\cite{simplebev} concatenates camera BEV feature map with rasterized Radar BEV feature map. FUTR3D~\cite{futr3d} fuses all the Radar, LiDAR and camera information with query-based Modality-Agnostic Feature Sampler (MAFS).

\subsection{Camera Stream Processing}

To keep the identical feature format, the BEV feature map is utilized in this research for LiDAR-camera fusion. LiDAR point cloud has precise spatial coordinates, it can be easily transferred to BEV feature representation with voxel-based encoders~\cite{pointpillars,voxelnet}. However, the images captured by cameras are in perspective-view and need more processing for multi-view images fusion and BEV projection. In camera stream processing, a 2D CNN backbone such as ResNet~\cite{resnet} or EfficientNet~\cite{efficientnet} is utilized to first extract the features from the input images. Then, the multi-view feature maps are projected to BEV with the cameras' intrinsics and extrinsics. It can be summarized as $\mathbf{F}=\mathbf{P}(I_1,I_2,\ldots,I_k)$ where existing projector $\mathbf{P}$ can project the perspective-view input feature maps $(I_1,I_2,\ldots,I_k)$ to a representation $\mathbf{F} \in \mathbb{R}^{H \times W \times C}$ in the BEV plane. Prior works differed in and contributed mainly toward the algorithms used for projecting features from 2D perspective-view to BEV. The projectors typically apply geometry-based and transformer-based approaches.

Geometry-based approaches project the perspective-view image features into BEV by using the geometric relationships. Lift-Splat~\cite{lss} lifts each 2D image to a frustum-shaped point cloud by predicting a categorical distribution over depth and a context vector. Then, the cameras' extrinsics and intrinsics are used to splat each frustum onto the BEV plane. In contrast to projecting the multi-view images into BEV, Simple-BEV~\cite{simplebev} defines a 3D volume over the BEV plane to project image features by bilinear sampling.

Transformer-based fusion leverages transformers to project multi-view images to BEV representation. Current works~\cite{bevformer,cobevt,cvt} define a BEV query and use cross-attention to link a BEV embedding to the multi-view images. CVT\cite{cvt} and CoBEVT~\cite{cobevt} adopt positional embedding in cross-attention to use the geometric cues of the cameras. BEVFormer~\cite{bevformer} proposes temporal self-attention to utilize the temporal cues by incorporating historical BEV information with current environment.

\begin{figure*}[htbp]
\centering
\includegraphics[width=0.99\textwidth]{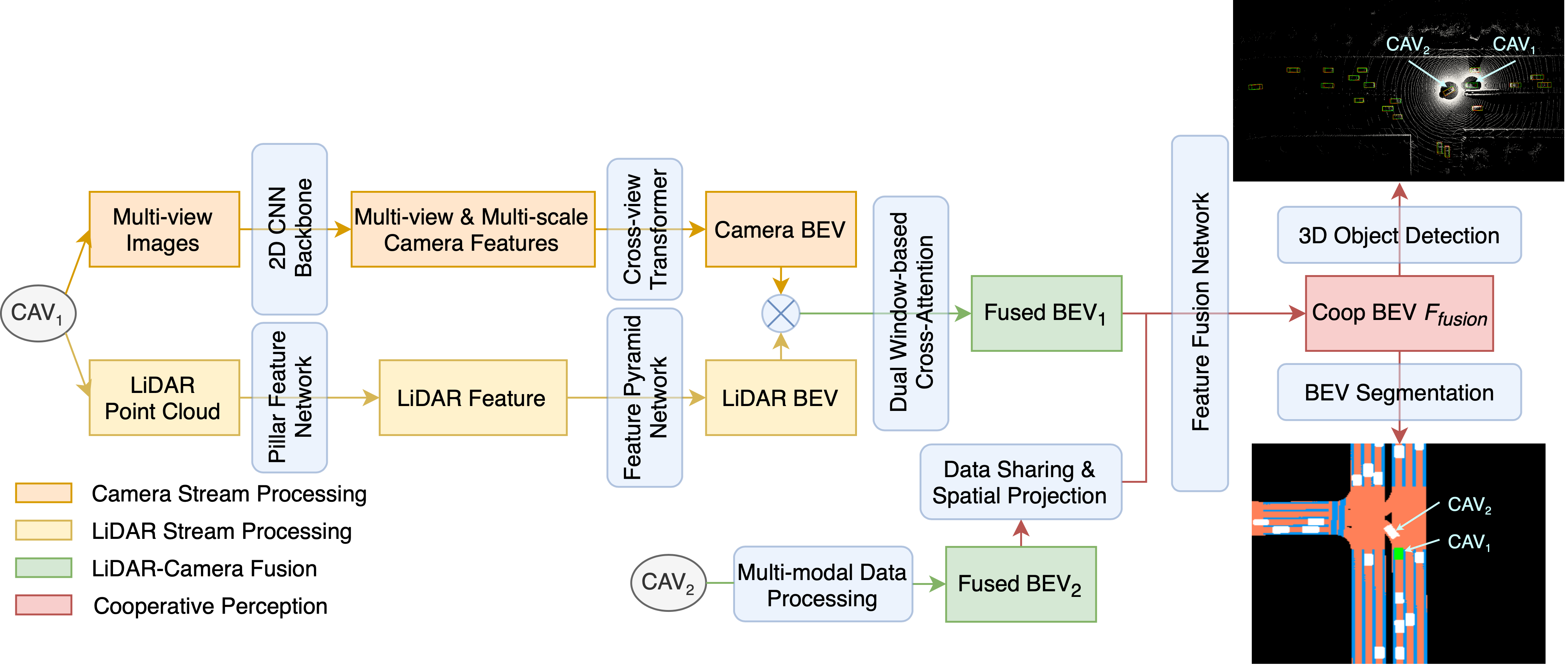}
\caption{Architecture of the Cooperative Perception framework with LiDAR-Camera Bird's-Eye View Fusion (CoBEVFusion). This figure depicts the data processing of two CAVs ($CAV_1$ and $CAV_2$). The architecture consists of five parts: camera stream processing (orange branch), LiDAR stream processing (yellow branch), LiDAR-camera fusion (green branch), cooperative feature fusion (red branch), and final perception head including BEV semantic segmentation and 3D object detection. The perception results are based on $CAV_1$'s view.}
\label{fig:cobevfusion}
\end{figure*}

\subsection{Cooperative Perception}

Cooper~\cite{cooper} utilizes early fusion by broadcasting raw LiDAR data which incurs highest data transfer cost. Late fusion~\cite{trupercept,dairv2x,zhang2021distributed} shares and aggregates the predictions from the CAVs to reduce data transfer burden. The contextual information gets lost in late fusion and the performance of cooperative perception highly relies on the other CAVs' individual prediction accuracy and the post-processing of the predictions. The performance of cooperative perception with early and late fusion can be improved by optimizing the 3D object detectors and post-processing method~\cite{zhang2021distributed}. 

In intermediate fusion, the CAVs process the traffic information gathered by multi-modal sensors locally, and then share the extracted intermediate feature maps with other CAVs by using the VC systems within the communication range. The receivers' receive processed traffic information from other CAVs which are at different locations. Therefore, accurate and optimized integration and processing of the information obtained from different locations is critical for effective intermediate feature fusion to enable accurate object detection. The maximum and summation are calculated at the overlaps of the intermediate features in~\cite{fcooper} and~\cite{sumcooper} respectively. In V2VNet~\cite{v2vnet}, a Graph Neural Network (GNN) is applied to represent a map of CAVs based on the geological coordinates to facilitate data fusion. Xu et al.~\cite{opv2v} propose AttFuse and leverage self-attention to fuse the intermediate feature maps. Transformers are utilized in V2X-ViT~\cite{v2xvit} and CoBEVT~\cite{cobevt} for cooperative perception with intermediate feature fusion. DiscoNet~\cite{disconet} constrains all the intermediate feature maps in the student model to match the correspondences in the teacher model through knowledge distillation, resulting in a collaborative student model. Qiao et al.~\cite{qiao2023adaptive} propose a lightweight adaptive feature fusion approach that adaptively selects spatial or channel features for information aggregation.
\section{Methodology}
\label{sec:methodology}

The overall architecture of our CoBEVFusion is illustrated in this section and the model architecture is shown in Fig.~\ref{fig:cobevfusion}. It can be split into five modules: a) LiDAR stream processing encodes LiDAR point cloud and generates LiDAR BEV representation; b) camera stream processing extracts features from the multi-view images and projects the features into BEV; c) LiDAR-camera BEV feature fusion aggregates LiDAR and camera BEV representations; d) cooperative feature fusion projects the fused BEV representations into receivers' coordinate systems and fuses the feature maps; and e) perception head predicts BEV semantic segmentation or detects objects. Fig.~\ref{fig:cobevfusion} depicts the data processing and interaction between two CAVs. The LiDAR stream processing, image stream processing and LiDAR-camera fusion are conducted locally. Then, the fused BEV representations are disseminated among the CAVs for cooperative perception.

\subsection{LiDAR Stream Processing}

The input point cloud with dimension $(n\times4)$ consists of $n$ points. Each point has attributes $(x, y, z)$ coordinates and intensity. Using the geological information, the LiDAR data can be encoded into BEV perspective easily. The Pillar Feature Network~\cite{pointpillars} is utilized to project the point cloud into 2D BEV pseudo image. 

First, the point cloud is partitioned into vertical columns (pillars) and PointNet~\cite{pointnet} encodes the pillars into 1D vectors. Then, the processed pillars are scattered back to the original locations and generate a 2D pseudo image in BEV. Feature Pyramid Network (FPN)~\cite{fpn} is utilized for further feature extraction and a 2D CNN layer decreases the number of channel to reduce the computational complexity for further feature fusion. Finally, the LiDAR point cloud is processed into the BEV representation $F_{LiDAR} \in \mathbb{R}^{H \times W \times C}$.

\subsection{Camera Stream Processing}

To perceive the surrounding traffic environment, the camera system of the AV utilizes $n$ cameras and generate $n$ RGB images in monocular views. The monocular views $(I_k, K_k, R_k, t_k)_{k=1}^n$ consist of input images $I_k \in \mathbb{R}^{h \times w \times c}$, camera intrinsic $K_k \in \mathbb{R}^{3 \times 3}$, rotation extrinsic $R_k \in \mathbb{R}^{3 \times 3}$, and translation $t_k \in \mathbb{R}^3$. First, a CNN backbone network processes the images and extracts multi-scale feature representations of the multi-view images. Then, Cross-view Transformer (CVT)~\cite{cvt} is utilized to project the image features into BEV. Finally, the camera BEV representation is upsampled to $F_{camera} \in \mathbb{R}^{H \times W \times C}$ by using 2D CNN layers and Bi-linear interpolation. 

\subsection{Bird's-Eye-View Fusion}

\begin{figure}[t]
\centering
    % (1)
    \begin{subfigure}{0.4\columnwidth}
        \includegraphics[width=\columnwidth]{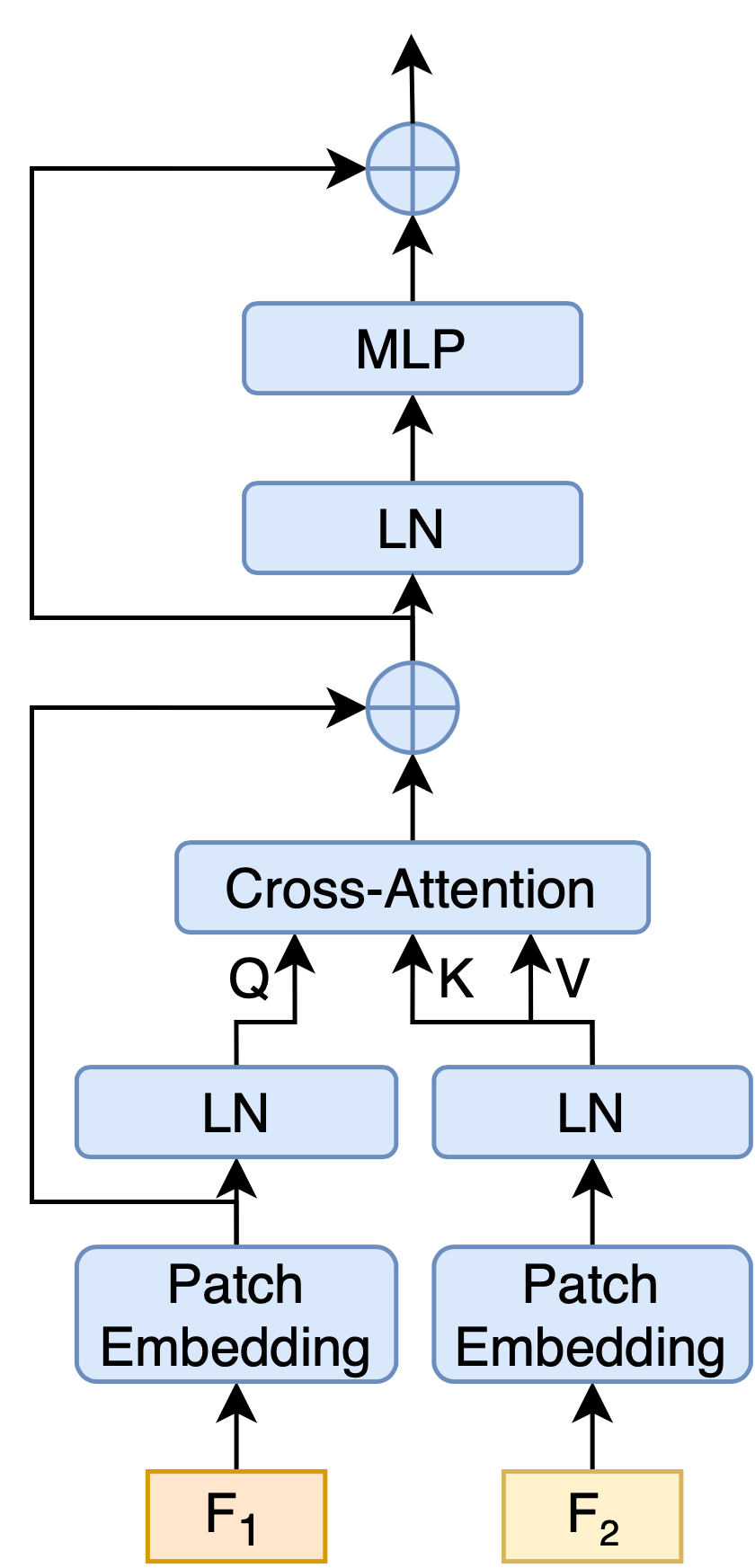}
        \caption{Window-based Cross-Attention (WCA)}
        \label{fig:wca}
    \end{subfigure}
    \hfill
    % (2)
    \begin{subfigure}{0.5\columnwidth}
        \includegraphics[width=\columnwidth]{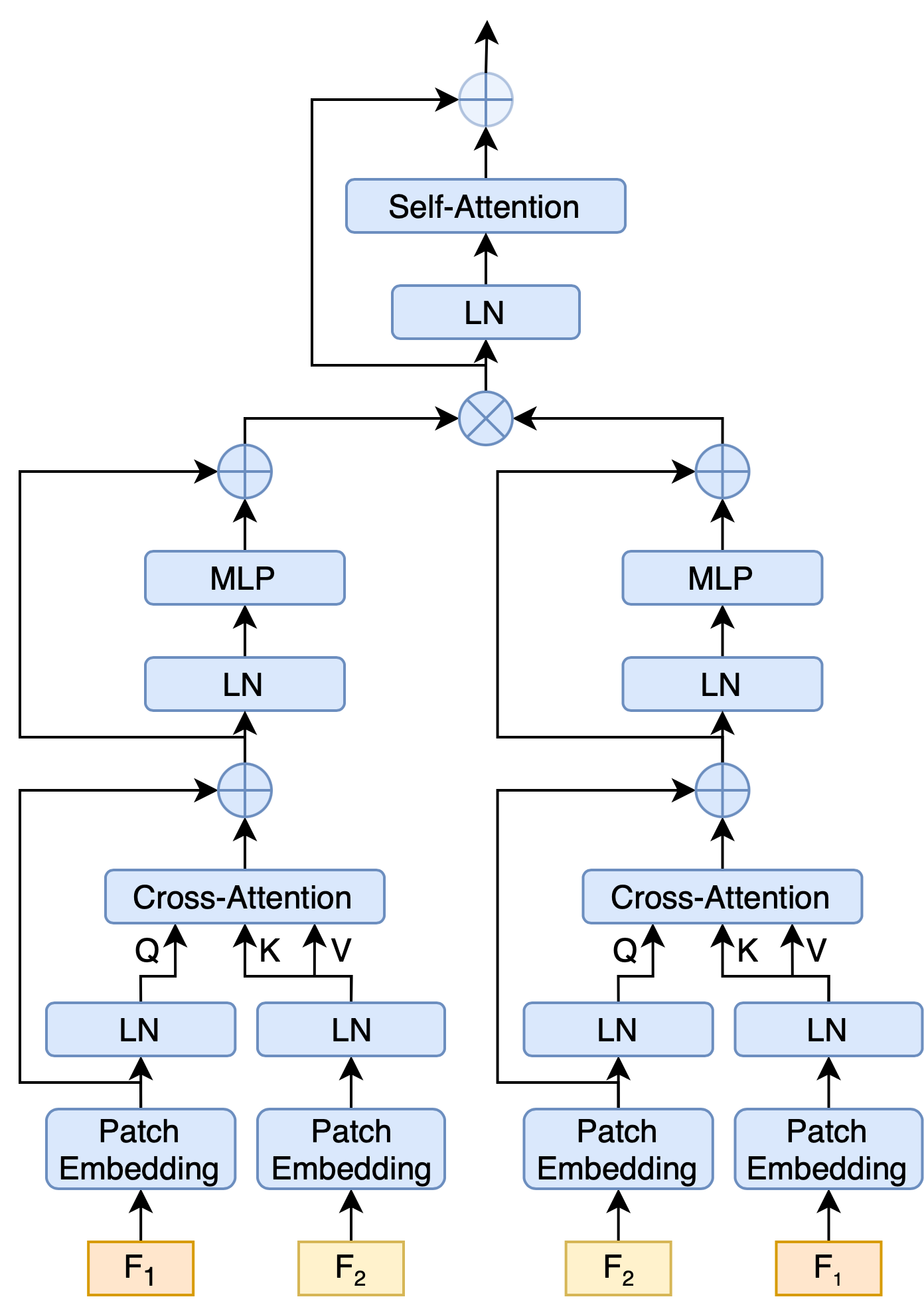}
        \caption{Dual Window-based Cross-Attention (DWCA).}
        \label{fig:dwca}
    \end{subfigure}
%\vspace*{-5mm}
\caption{Architecture of LiDAR-camera fusion model with cross-attention.}
\label{fig:ca}
\end{figure}

The LiDAR and camera BEV representations are fused together locally before data dissemination. We propose a Dual Window-based Cross-Attention (DWCA) model for LiDAR-camera fusion. The architecture of DWCA is shown in Fig.~\ref{fig:dwca} which contains two WCA (Fig.~\ref{fig:wca}) modules and one self-attention module. As a consequence of the aggregated output from WCA, the perception primarily relies on one kind of feature in the next step. DeepFusion~\cite{deepfusion} concatenates the original LiDAR feature with the aggregated camera feature for 3D object detection, whereas the two concatenated features are not equivalent. As for semantic segmentation, the texture information of cameras is critical for lane and drivable area detection. Therefore, we utilize two WCA modules with reversed inputs for LiDAR-camera fusion.

% \begin{figure}[t]
% \centering
% \includegraphics[width=0.85\columnwidth]{figures/dwca.png}
% \caption{Dual Window-based Cross-Attention (DWCA) for feature fusion.}
% \label{fig:dwca}
% \end{figure}

On the left WCA branch of Fig.~\ref{fig:dwca}, the inputs $F_1$ and $F_2 \in \mathbb{R}^{H \times W \times C}$ are patch embedded into $(\frac{H}{P_1}\times \frac{W}{P_2}, P_1\times P_2, C)$ by using a window of size $(P_1, P_2)$. The embedded features are layer normalized and transformed to query $Q_{F_1}$, key $K_{F_2}$, and value $V_{F_2}$ by using three linear networks. The cross-attention is calculated with Eq.\ref{eq:ca}. We compute the dot product between the query of $F_1$ and key of $F_2$, and divide it by a scale factor $\sqrt{d_k}$. The softmax function is used to calculate the attention weights, and the final cross-attention representation is calculated by multiplying the attention weights by the value of $F_2$. Layer normalization and a linear layer with residual skip connection are utilized to calculate the final output. The right WCA branch in Fig.~\ref{fig:dwca} has the same architecture as the left one, but with reversed inputs. These two WCA blocks make the LiDAR and camera features reinforce each other. The outputs from the two WCAs are concatenated together and a self-attention layer is utilized to further align the features and generate the fused LiDAR-camera representation.

\begin{equation}
CA(Q_{F_1}, K_{F_2}, V_{F_2}) = softmax(\frac{Q_{F_1}K_{F_2}^T}{\sqrt{d_k}})V_{F_2}
\label{eq:ca}
\end{equation}

% (B, C, H, W) -> (Bxhxw, p1xp2, C) -> to-qkv -> (Bxhxw, p1xp2, dhxh)
% (Bxhxw, p1xp2, dhxh) -> (Bxhxw, h, p1xp2, dh)

% $F\!\mbox{-}\mathrm{an}$
% $F'=W\!\mbox{-}CA(LN(F_1), LN(F_2))+F_1$
% $F_{fuse}=MLP(LN(F'))+F'$
% \noindent where the $W\!\mbox{-}CA$ is computed with following equation:

% \begin{figure}[htbp]
% \centering
% \includegraphics[width=0.95\columnwidth]{figures/covit.png}
% \caption{Cooperative representation fusion with CrossViT.}
% \label{fig:covit}
% \end{figure}

%\textcolor{mygreen}{Green} branch is LC-Coop and  \textcolor{myred}{red} branch is Coop-LC.

\subsection{Cooperative Feature Fusion}

The above mentioned of LiDAR stream processing, camera stream processing and LiDAR-camera fusion are all completed in the perspective CAVs' coordinate systems. The feature representations are broadcasted to other CAVs and projected into receivers' coordinate systems based on the geological information. After feature projection, a 3D convolutional neural network as described in~\cite{qiao2023adaptive} is utilized to aggregate the feature representations from multiple CAVs.

\subsection{Perception Head}

\subsubsection{BEV Semantic Segmentation}
A 2D CNN layer is used to generate the final segmentation output. The weighted cross entropy loss is used to train the semantic segmentation model.

\subsubsection{3D Object Detection}

The fused features $F_{fusion}$ are fed into a SSD~\cite{ssd} that can predict the confidence scores for the detected object classes and regress the 3D bounding boxes. The loss function consists of focal loss~\cite{retinanet} for classification, and smooth $L_1$ loss for regression. The complete loss function of the detection model is given below:
\begin{equation}
\begin{split}
L &= \beta_{cls}L_{cls} + \beta_{reg}L_{reg}\\
     &= \beta_{cls}L_{focal}(p) + \beta_{reg}smooth_{L_{1}}(sin(q-y_{reg}))
\end{split}
\label{eq:loss}
\end{equation}

\noindent where $\beta_{cls}$ and $\beta_{reg}$ are the classification loss and regression loss coefficients respectively, $p$ is the prediction probability, $q$ is the number of anchor boxes and $y_{reg}$ is the number of ground truth boxes.

\section{Experiments}
\label{sec:experiments}

We conduct experiments on the publicly available cooperative perception datasets OPV2V dataset~\cite{opv2v}. The LiDAR-camera fusion model is evaluated with both single vehicle perception and cooperative perception on two perception tasks including BEV semantic segmentation and 3D object detection. We compare the predicted results with conventional single vehicle perception model (no fusion) and multiple SOTA cooperative perception models.

\subsection{Datasets}

The OPV2V dataset is built with OpenCDA simulation tool~\cite{opencda} and includes two subsets, a default CARLA towns and a Culver City. The default CARLA towns contains 6,765 training samples, 1,980 validation samples, and 2,170 testing samples in eight CARLA default towns. The Culver City contains 550 samples to test the domain adaptation ability of the model. The number of CAVs in this dataset ranges between [2, 7], and each CAV has its own LiDAR information, four cameras' data, labeled 3D bounding boxes, and BEV semantic segmentation ground truth. The BEV semantic segmentation has four classes including background, vehicle, drivable area and lane.

\subsection{Implementation Details}

During training, a random group of CAVs are selected from the scene with a defined upper limit of CAVs including an ego vehicle. For validation purposes, the ego vehicle and the CAVs are fixed for a fair comparison. Our model is implemented using the PyTorch framework, trained and evaluated on NVIDIA A100 80GB GPU. Early stopping, multi-step scheduler, and AdamW optimizer with an $\epsilon$ of 0.1 and a weight decay of 0.0001 are used to train the network. To compare with other benchmarks, we follow the parameter settings in~\cite{cobevt} for BEV semantic segmentation and~\cite{opv2v} for 3D object detection. 

The images are resized into $512 \times 512$ and $512 \times 640$ in BEV segmentation and 3D object detection respectively. The perception ranges of $x, y$ of these two tasks are [(-50, 50), (-50, 50)] and [(-140, 140), (-40, 40)] meters. The size of the vehicle anchor for object detection has a (length, width, height) of (3.9, 1.6, 1.56) meters.

\begin{figure}[t]
\centering
\includegraphics[width=0.99\columnwidth]{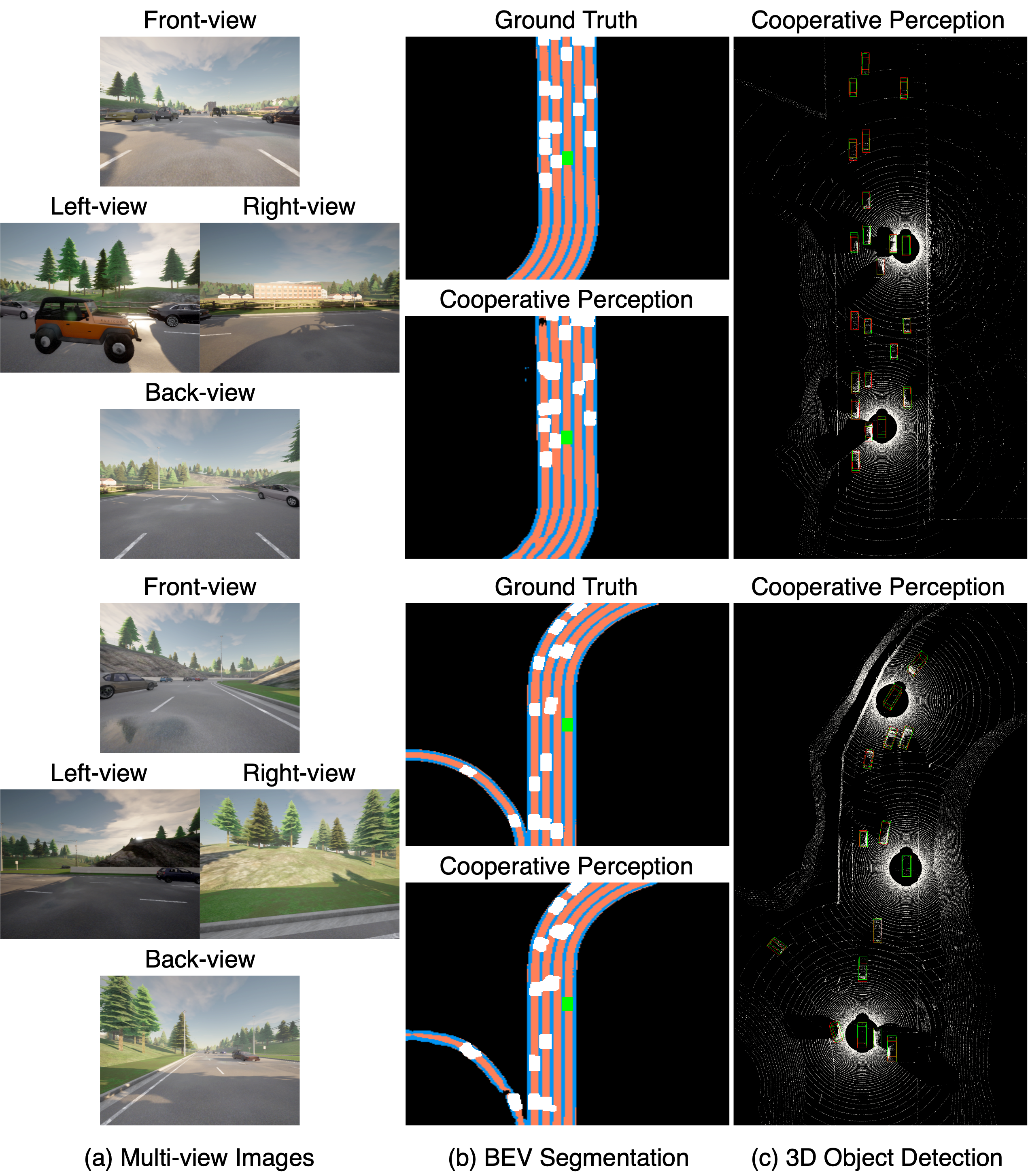}
\caption{Qualitative results of cooperative perception on OPV2V dataset for BEV semantic segmentation and 3D object detection. (a) Multi-view images including front-view, back-view, left-view and right-view. (b) BEV semantic segmentation ground truths and prediction results. (c) 3D object detection results.}
\label{fig:results}
\end{figure}

% \begin{figure*}
% \centering
% \includegraphics[width=0.99\textwidth]{figures/example.png}\\
% \includegraphics[width=0.99\textwidth]{figures/example.png}\\
% \includegraphics[width=0.99\textwidth]{figures/example.png}\\
% \caption{Qualitative results of cooperative perception on OPV2V dataset for BEV semantic segmentation and 3D object detection.}
% \label{fig:qualitative_results}
% \end{figure*}

\subsection{Results and Discussion}

We first compare our LiDAR-camera fusion model DWCA with perception models with single-modal data and other SOTA LiDAR-camera fusion models~\cite{bevfusion2,bevfusion1} under single vehicle perception mode. After that, we compare our CoBEVFusion with other cooperative perception SOTA models. The evaluation results for BEV semantic segmentation and 3D object detection on OPV2V test set are listed in Table~\ref{table:opv2v_bev} and~\ref{table:opv2v_3d} respectively.

Two qualitative results of our CoBEVFusion are shown in Fig.~\ref{fig:results}. Each case displays the four-view camera images, the BEV semantic segmentation ground truth and prediction, and visualization of 3D object detection results on LiDAR point cloud.

% {m{0.05\columnwidth}|m{0.3\columnwidth}|m{0.1\columnwidth}m{0.1\columnwidth}|m{0.1\columnwidth}m{0.1\columnwidth}||m{0.1\columnwidth}m{0.1\columnwidth}|m{0.15\columnwidth}||m{0.15\columnwidth}}
% \hline

% \multirow{3}{*}{} &
% \multirow{3}{*}{\textbf{Method}} & \multicolumn{4}{c||}{\textbf{OPV2V}} & \multicolumn{3}{c||}{\textbf{CODD}} & \\

\begin{table}[t]
\caption{Evaluation results on the OPV2V test set for Bird's-Eye-View (BEV) segmentation.}
\centering
% \begin{tabular}{m{0.1\textwidth}|m{0.05\textwidth}|m{0.05\textwidth}|m{0.05\textwidth}|m{0.05\textwidth}}
\begin{tabular}{l|c|ccc}
\hline

\multirow{2}{*}{\textbf{Method}} & \multirow{2}{*}{\textbf{M.}} & \textbf{Vehicle} & \textbf{Dr.Area} & \textbf{Lane}\\
 &  & mIoU & mIoU & mIoU\\
\hline
\hline
\multicolumn{5}{c}{\textbf{Single Vehicle Perception (No Fusion)}}\\
\hline
\hline
PointPillars~\cite{pointpillars} & L & 31.9 & 51.8 & 38.0\\
CVT~\cite{cvt} & C & 37.7 & 57.8 & 43.7\\
SinBEVT~\cite{cobevt} & C & 38.8 & 59.7 & 47.0\\
BEVFusion~\cite{bevfusion1} & LC & 38.2 & 59.9 & 46.9\\
BEVFusion~\cite{bevfusion2} & LC & 36.6 & 61.0 & 46.4\\
\hline
\textbf{DWCA} & LC & \textbf{40.4} & \textbf{61.4} & \textbf{47.6}\\
\hline
\hline
\multicolumn{5}{c}{\textbf{Cooperative Perception}}\\
\hline
\hline
F-Cooper~\cite{fcooper} & C & 52.5 & 60.4 & 46.5\\
V2VNet~\cite{v2vnet} & C & 53.5 & 60.2 & 47.5\\
%\hline
AttFuse~\cite{opv2v} & C & 51.9 & 60.5 & 46.2\\
DiscoNet~\cite{disconet} & C & 52.9 & 60.7 & 45.8\\
FuseBEVT~\cite{cobevt} & C & 59.0 & 62.1 & 49.2\\
CoBEVT~\cite{cobevt} & C & \textbf{60.4} & \textbf{63.0} & \textbf{53.0}\\
\hline
% CoBEVT~\cite{cobevt} & C & 56.8 & \textbf{63.0} & \textbf{53.0}\\
\textbf{CoBEVFusion} & LC & 59.5 & 61.7 & 50.9\\
\hline
\end{tabular}
\\
\footnotesize{\raggedleft M.: Modality. Dr.Area: Drivable area. \par}
\label{table:opv2v_bev}
\end{table}

\subsubsection{BEV Semantic Segmentation}

In single vehicle BEV semantic segmentation, our LiDAR-camera fusion module DWCA achieves highest mIoU on vehicle, drivable area, and lane segmentation at 40.4\%, 61.4\% and 47.6\% which are 1.6\%, 0.4\% and 0.6\% higher than other single vehicle perception models. 

\begin{figure*}[t]
\centering
\includegraphics[width=0.8\textwidth]{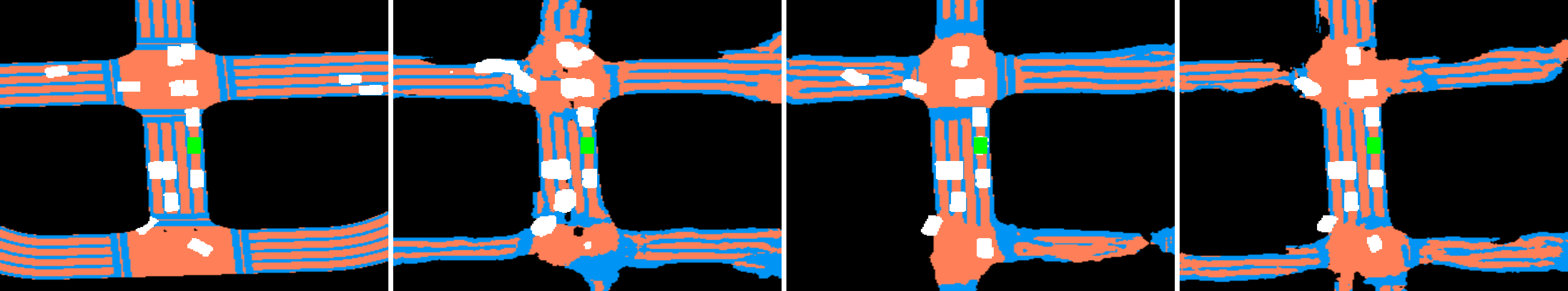}
\includegraphics[width=0.8\textwidth]{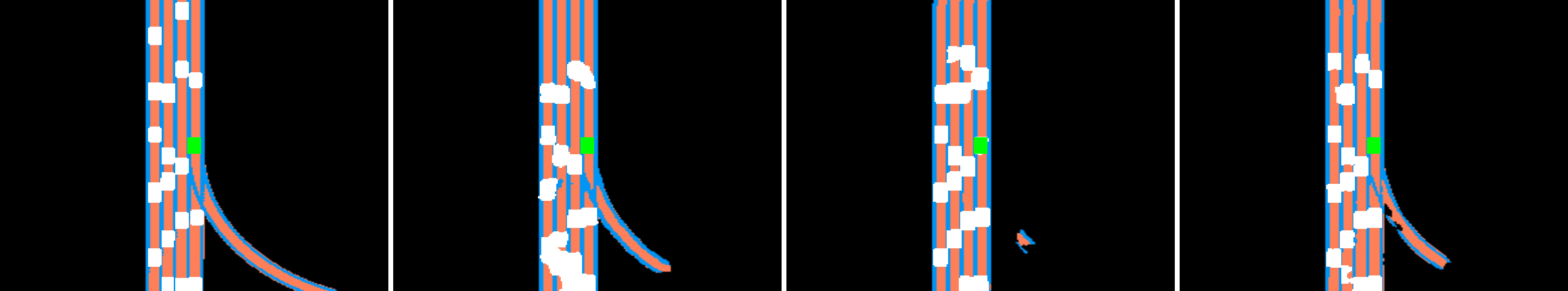}
\includegraphics[width=0.8\textwidth]{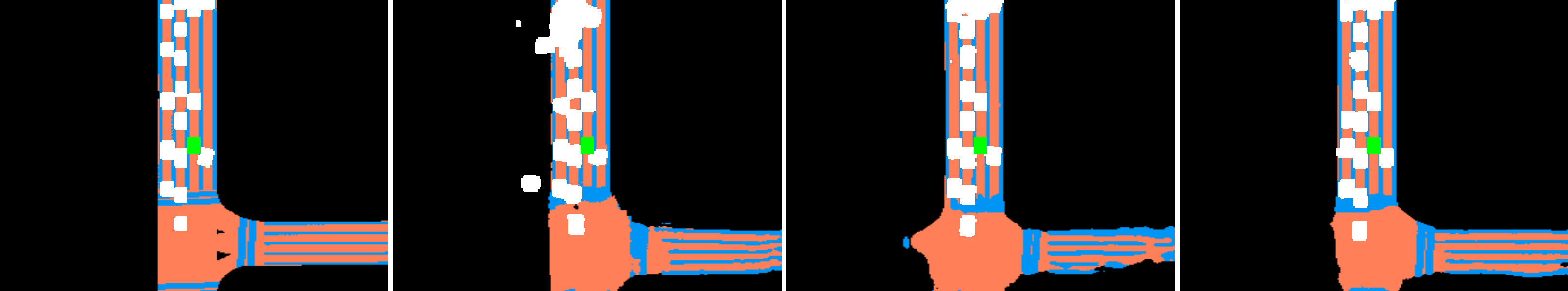}
\includegraphics[width=0.8\textwidth]{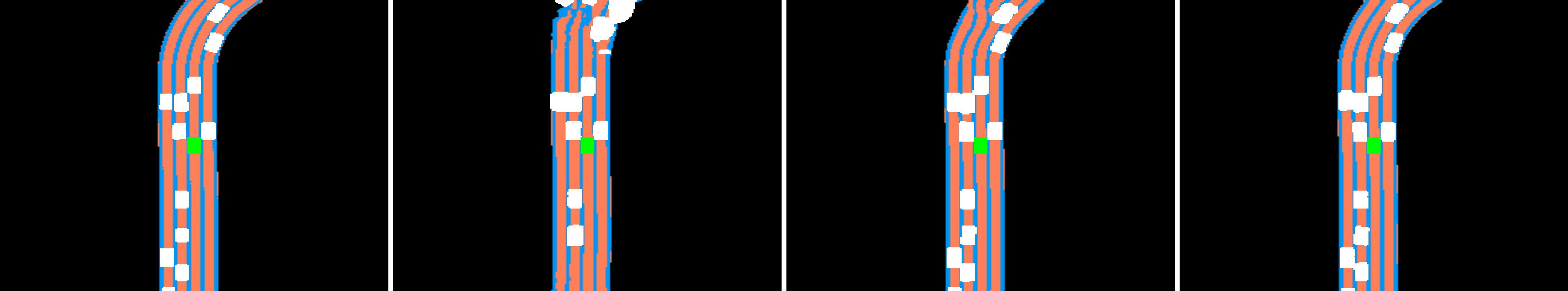}
\includegraphics[width=0.8\textwidth]{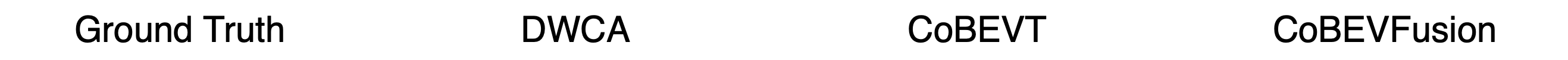}
\caption{Qualitative results on OPV2V dataset for BEV semantic segmentation with cooperative perception.}
\label{fig:segmentation}
\end{figure*}

Our CoBEVFusion surpasses single-vehicle perception models and most of other SOTA cooperative perception models. It also achieves comparable results with the best camera-only model CoBEVT~\cite{cobevt}. Some qualitative results of the BEV segmentation are shown in Fig.~\ref{fig:segmentation} for comparison. Involving the LiDAR for semantic segmentation effects the inference on some details, but it extends the perception field and resolution distance in some cases. Meanwhile, the cooperative perception brings more information for the ego vehicle, which makes the inference on further objects more accurate. 

\subsubsection{3D Object Detection}

Table~\ref{table:opv2v_3d} displays the evaluation results on OPV2V Default CARLA Towns set and Culver City set for vehicle detection and domain adaptation respectively. Our DWCA surpasses the LiDAR-based 3D object detection model PointPillars with large margin. Our LiDAR-camera fusion-based cooperative perception model, CoBEVFusion, outperforms all the other models by at least 0.9\% on vehicle detection and 0.4\% on domain adaptation.

\begin{figure*}[t]
\centering
\includegraphics[width=0.9\textwidth]{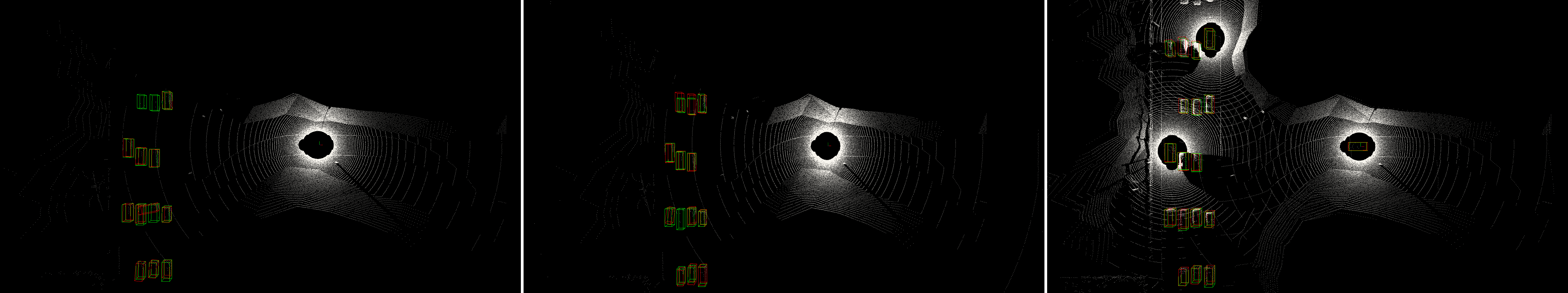}
\includegraphics[width=0.9\textwidth]{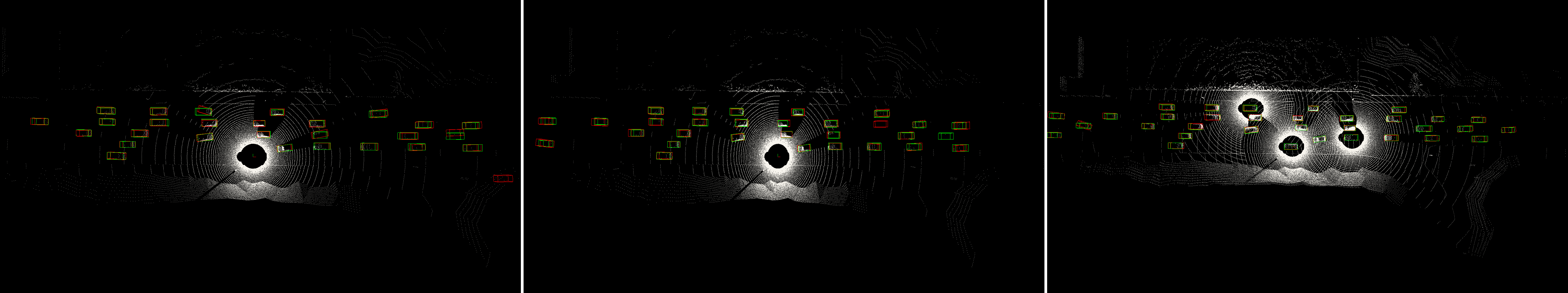}
\includegraphics[width=0.9\textwidth]{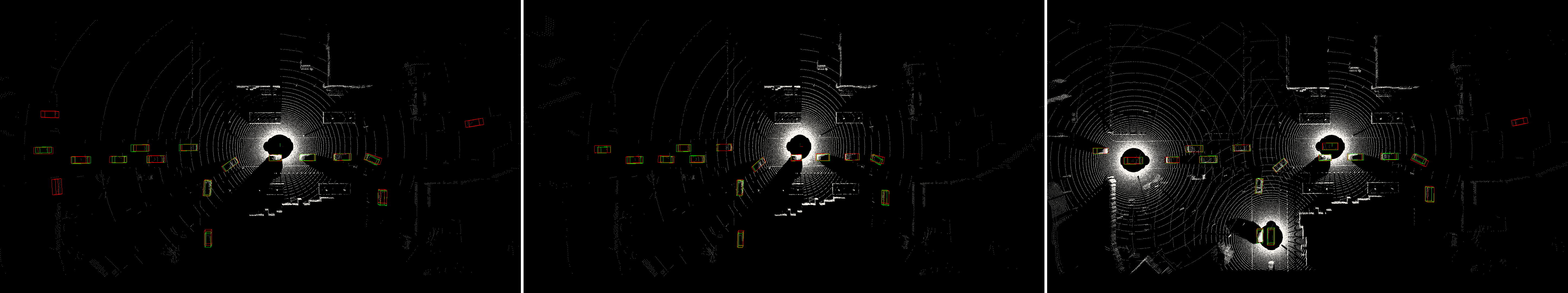}
\includegraphics[width=0.8\textwidth]{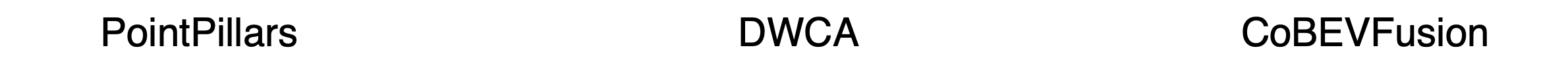}
\caption{Qualitative results on OPV2V dataset for 3D object detection with cooperative perception. }
\label{fig:detection}
\end{figure*}

In the 3D object detection qualitative results Fig.~\ref{fig:detection}, the cooperative perception improves the detection of distant vehicles significantly. Also, the cooperative perception helps the ego vehicles to have a better and earlier understanding of the traffic environment before passing the intersections as shown in the third case of qualitative results.

\subsection{Ablation Study}

To prove the effectiveness of our fusing strategy for two data modalities, we conduct ablation experiments on WCA with reversed inputs for single vehilce perception. The ablation study results is shown in Table~\ref{table:cobevfusion_ablation}. The baseline is perception model with single-modal data including camera-based BEV semantic segmentation with CVT~\cite{cvt} and LiDAR-based 3D object detection with PointPillars~\cite{pointpillars}. WCA (LC) represents $F_1$ and $F_2$ are LiDAR and camera features respectively, while WCA (CL) inputs reversed features with WCA (LC). The experiments illustrate that when the query vector is LiDAR stream, the model achieves better performance on vehicle perception such as vehicle segmentation and 3D vehicle detection. However, when the query vector is the camera stream, the model achieves higher mIoU in drivable area and lane segmentation. DWCA, a combination of these two WCA blocks, outperforms on both BEV semantic segmentation and 3D object detection.

\begin{table}[t]
\caption{Evaluation results on the OPV2V datsets including Default CARLA Towns test set for vehicle detection and Culver City for domain adaptation.}
\centering
%\begin{tabular}{m{0.35\columnwidth}|m{0.1\columnwidth}m{0.1\columnwidth}|m{0.1\columnwidth}m{0.1\columnwidth}}
\begin{tabular}{l|c|ccc}
\hline
\multirow{2}{*}{\textbf{Method}} & \multirow{2}{*}{\textbf{M.}} & \multicolumn{2}{c}{\textbf{Default}} & \textbf{Culver}\\
 &  & AP@.5 & AP@.7 & AP@.7\\
\hline
\hline
\multicolumn{5}{c}{\textbf{Single Vehicle Perception (No Fusion)}}\\
\hline
\hline
PointPillars~\cite{pointpillars} & L & 67.9 & 60.2 & 47.1\\
\hline
\textbf{DWCA} & LC & 83.4 & 64.7 & 56.5\\
\hline
\hline
\multicolumn{5}{c}{\textbf{Cooperative Perception}}\\
\hline
\hline
Early Fusion & L & 89.1 & 80.0 & 69.6\\
Late Fusion & L & 85.8 & 78.1 & 66.8\\
\hline
AvgFusion~\cite{qiao2023adaptive} & L & 84.3 & 74.7 & 68.0\\
F-Cooper~\cite{fcooper} & L & 88.7 & 79.0 & 72.8\\
V2VNet~\cite{v2vnet} & L & 89.7 & 82.2 & 73.4\\
AttFuse~\cite{opv2v} & L & 90.8 & 81.5 & 73.5\\
V2X-ViT~\cite{v2xvit} & L & 89.1 & 82.6 & 73.7\\
FuseBEVT~\cite{cobevt} & L & - & 85.2 & 74.6\\
C-3DFusion~\cite{qiao2023adaptive} & L & 90.8 & 83.6 & 75.7\\
C-AdaFusion~\cite{qiao2023adaptive} & L & 88.5 & 81.4 & 72.4\\
S-AdaFusion~\cite{qiao2023adaptive} & L & 91.6 & 85.6 & 79.0\\
\hline
\textbf{CoBEVFusion} & LC & \textbf{92.5} & \textbf{86.5} & \textbf{79.4}\\
\hline
\end{tabular}
\\
\footnotesize{\raggedleft M.: Modality. \par}
\label{table:opv2v_3d}
\end{table}

\begin{table}[t]
\caption{Evaluation results on the OPV2V datset for ablation study.}
\centering
\begin{tabular}{l|cccc}
\hline
\multirow{2}{*}{\textbf{Method}} & \textbf{Dr.Area} & \textbf{Lane} & \textbf{Vehicle} & \textbf{Default}\\
 & mIoU & mIoU & mIoU & AP@.7\\
\hline
Baseline & 57.8 & 43.7 & 37.7 & 60.2\\
WCA (LC) & 58.3 & 43.3 & 39.5 & 63.9\\
WCA (CL) & 60.2 & 46.5 & 35.4 & 62.5\\
\textbf{DWCA} & \textbf{61.4} & \textbf{47.6} & \textbf{40.4} & \textbf{64.7}\\
% WCA (LC) & 39.5 & 58.3 & 43.3 & 63.9 & 57.8\\
% WCA (CL) & 35.4 & 60.2 & 46.5 & 62.5 & 52.1\\
% DWCA & 40.4 & 61.4 & 47.6 & 64.7 & 56.5\\
\hline
\end{tabular}
\\
\footnotesize{\raggedleft Dr.Area: Drivable area. \par}
\label{table:cobevfusion_ablation}
\end{table}

\section{Conclusion}
\label{sec:conclusion}

In order to enhance vehicle perception, multi-modal inputs are required for cost-effective communication among CAVs as well as reliable data fusion. In this paper, we research on utilizing the multi-modal LiDAR-camera fusion feature for cooperative perception. We propose a Dual Window-based Cross-Attention (DWCA) model for LiDAR-camera BEV fusion. The fused BEV representation is utilized in cooperative perception to enhance the perception accuracy. The model is evaluated on a large scale cooperative perception benchmark dataset, OPV2V~\cite{opv2v}, for BEV semantic segmentation task and 3D object detection task. The proposed LiDAR-camera fusion model outperforms the perception models with single-modal data and other SOTA BEV fusion models. Our cooperative perception architecture also achieves SOTA performance in 3D object detection. 

From the visualized prediction results, we found that in some cases, the feature projection and fusion of cooperative perception models reduced the accuracy of referencing nearby targets. Additionally, OPV2V is a simulated dataset and the performance of cooperative perception requires to be evaluated in real-world. The latency and quality of data transfer must also be explored further in real-world environment for advancement in cooperative perception. 

{
    \small
    \bibliographystyle{ieeenat_fullname}
    \bibliography{main}
}


\end{document}